\RequirePackage{fix-cm}
\documentclass[twocolumn,bibtex]{svjour3}          
\smartqed  
\usepackage{graphicx}
\usepackage{amsmath}
\usepackage{mathptmx}
\usepackage{enumerate}
\usepackage[misc]{ifsym}
\usepackage[english]{babel}
\usepackage[utf8]{inputenc}
\usepackage[T1]{fontenc}
\usepackage{latexsym}
\usepackage{multicol}
\usepackage{sidecap}
\usepackage{wrapfig}
\usepackage{graphicx}
\usepackage{floatrow}
\usepackage{graphicx}
\usepackage{sidecap}
\usepackage{xcolor}
\usepackage{tabularx}
\usepackage{multirow}
\usepackage{url}
\usepackage{hyperref}
\hypersetup{colorlinks=true, linkcolor=blue, filecolor=blue, urlcolor=blue, citecolor=blue}
\usepackage{adjustbox}
\usepackage{ulem}

\begin{document}

\title{Deep learning for lithological classification of carbonate rock micro-CT images}

\author{Carlos E. M. dos Anjos\textsuperscript{1} \and
        Manuel R. V. Avila\textsuperscript{1} \and
        Adna G. P. Vasconcelos\textsuperscript{1} \and
        Aurea M. P. Neta\textsuperscript{1} \and
        Lizianne C. Medeiros\textsuperscript{1} \and
        Alexandre G. Evsukoff\textsuperscript{1} \and
        Rodrigo Surmas\textsuperscript{2} \and
        Luiz Landau\textsuperscript{1}}

\institute{Carlos Menezes (\Letter) \at
              carlos.menezes@poli.ufrj.br
          \and
          \at
          $^1$\textit{Department of Civil Engineering, COPPE, Federal University of Rio de Janeiro, Brazil}\\
          $^2$\textit{CENPES, Petrobras}
}

\date{Received: date / Accepted: date}

\maketitle
\begin{abstract}
\sloppy
In addition to the ongoing development, pre-salt carbonate reservoir characterization remains a challenge, primarily due to inherent geological particularities. These challenges stimulate the use of well-established technologies, such as artificial intelligence algorithms, for image classification tasks. Therefore, this work intends to present an application of deep learning techniques to identify patterns in Brazilian pre-salt carbonate rock microtomographic images, thus making possible lithological classification. Four convolutional neural network models were proposed. The first model includes three convolutional layers followed by fully connected layers and is used as a base model for the following proposals. In the next two models, we replace the max pooling layer with a spatial pyramid pooling and a global average pooling layer. The last model uses a combination of spatial pyramid pooling followed by global average pooling in place of the last pooling layer. All models are compared using original images, when possible, as well as resized images. The dataset consists of 6,000 images from three different classes. The model performances were evaluated by each image individually, as well as by the most frequently predicted class for each sample. According to accuracy, Model 2 trained on resized images achieved the best results, reaching an average of 75.54\% for the first evaluation approach and an average of 81.33\% for the second. We developed a workflow to automate and accelerate the lithology classification of Brazilian pre-salt carbonate samples by categorizing microtomographic images using deep learning algorithms in a non-destructive way.

 \end{abstract}


\section{Introduction}
\label{sec:1}
\sloppy
Carbonate reservoir characterization involves a set of elaborated methods that help to understand the physical and geological processes that make up the system, comprising geologic, physical, chemical, and mathematical modeling techniques to describe the composition and arrangement of rock layers, as well as their physical properties \cite{buryakovskyetal}.

Three main properties are linked and closely related to reservoir quality and the viability of reservoir development: porosity, permeability, and lithology \cite{valentin2019deep}. Rock may serve as a reservoir if it includes void space to store commercial volumes of hydrocarbons and its porous system is interconnected and able to deliver the hydrocarbons to extraction wells. Both of those characteristics are linked to the lithology of a rock, which determines the exploitation reservoir feasibility \cite{valentin2019deep}. Accurate discrimination of lithology helps to reduce the error in the prediction of permeability and hydrocarbon volume as well as to understand the depositional and diagenetic processes, which are also closely correlated with implications for fluid flow properties \cite{ahr}.

Usually, two-dimensional thin-section images form the basis of lithology definition \cite{remeysenandswennen}. As noticed by Valentín et al. \cite{valentin2019deep}, this procedure is subjective once it depends on the chosen attributes used for the classification. For instance, this activity may focus on biological settings or emphasize petrological characteristics by accounting for the granulometry and mineralogy, for example. Very often, the amount of information to be taken into account in each case, when analyzing a particular reservoir, depends significantly on subjective intuition rather than on quantitative objective measurements \cite{tschannen}.

According to Ketcham and Carlson \cite{ketchamandcarlson}, data extracted from a three-dimensional rock model are closely related to those obtained in a more exhaustive procedure (i.e., thin-sections) and are effortless and natural to interpret. X-ray computed tomography (CT) is a popular tool to generate a three-dimensional models of a sample without destroying it. This technique was initially used for medical purposes in the 1970s, but in the following years it has been applied to investigate Earth materials such as those related to petroleum geology \cite{vinegar,honarpouretal,wildenschildandsheppard}. Currently, this technique is considered to be the most practical method to obtain the three-dimensional inner structure of porous media \cite{wangetal}. Additionally, once the data are digitized, it is straightforward to use the data to accomplish quantitative and qualitative analysis.

In this manner, micro-CT has become a standard technique in reservoir characterization workflows \cite{knackstedtetal,arnsetal,claesetal} because it allows for a representative description of microstructure and contributes to the understanding of the physical phenomena of fluid flow \cite{dvorkinetal2009,muljadietal} and estimation of mechanical properties \cite{arnsetal2002,knackstedtetal2009}.

In the same fashion as in the digital petrophysics field, the increasing computational capabilities also enabled the development of more sophisticated artificial intelligence algorithms, such as deep learning. Those algorithms have been demonstrating excellent performance in tasks such as automated quantification of lung cancer radiographic characteristics \cite{hosny2018deep} and hand bone age assessment for evaluation of endocrine and metabolic disorders \cite{lee2017fully}, among others, in the field of medicine. This success has led to a growing interest in using artificial intelligence techniques to automate some processes and as a decision support system in many industrial applications, thus allowing the experts to focus on the most complicated tasks that require human intervention. In the oil and gas industry, these techniques are useful in applications from the reservoir exploration stage to production when using different data domains, for example, seismic \cite{waldelandandsolberg,pochetetal} and petrophysical data \cite{saggafandnebrija,quiandcarr,maitietal,jafarietal}.

In the digital petrophysics field, Marmo et al. \cite{marmoetal} applied a multi-layer perceptron neural network to identify carbonate textures and recognize their original depositional environments in 8-bit linear gray-tone digital images of thin-sections; Cheng and Guo \cite{chengandguo} uses deep convolutional neural network (CNN) to identify sandstone granularity, with high reliability, in colored images of thin-sections; and de Lima et al. \cite{delimaetal} used transfer learning algorithms to classify uninterpreted images, such as microfossils, cores, or petrographic photomicrography, along with rock and mineral hand sample images.

With regard to the micro-CT image context, Bordignon et al. \cite{bordignonetal} propose a CNN to characterize the grain size distributions of porous rocks based only on a synthetic training dataset; Karimpouli and Tahmasebi \cite{karimpouliandtahmasebi} applies a CNN to estimate the P- and S-wave velocities and formation factors, without time-demanding numerical simulations; Karimpouli and Tahmasebi \cite{karimpouli2019segmentation} also proposed a method based on CNN to automate porous phase segmentation in a more accurate and reliable way; Sudakov et al. \cite{sudakov2019driving} employed CNN to estimate permeability values simulated with pore network approach; and Alqahtani et al. \cite{alqahtanietal} uses CNN to rapidly estimate porosity, specific surface area, and average pore size.

Several works pursuing lithofacies prediction have also been using artificial intelligence algorithms to reduce uncertainties in estimating hydrocarbon saturation through well log data \cite{chaietal,wangandcarr,kenari2013robust,amiri2015improving,almudhafar}. Odi and Nguyen \cite{odiandnguyen} developed a combined approach to predict geological facies by utilizing molecular weight, density, and porosity from CT images of core samples to learn from existing user-defined geological facies classification. 

On a smaller scale, this work aims to identify the lithology of rocks by using micro-CT images to establish exploration and development strategies for carbonate reservoirs. A mechanism to assist the specialist in the lithological classification of carbonate rock samples using micro-CT images, thus speeding reservoir evaluation, is shown here.

This work is organized into seven sections. The second denotes the conventional workflow for lithology identification through thin-sections, as well as our proposed method. The third synthesizes the theory behind CNN and its application to image analysis. The fourth section exposes the available dataset of micro-CT images of carbonate rock samples, the models used for the lithology classification task, how those models were trained, and how they were evaluated. The results are summarized in section five. The sixth section discusses the obtained results. Finally, in the last section, conclusions emphasize the positive view of using micro-CT images for lithologic classification based on CNN, and ongoing future works and proposals are stated.


\section{Conventional and proposed workflows}

\begin{figure*}[!ht]
    \centering
    \includegraphics[scale=0.3]{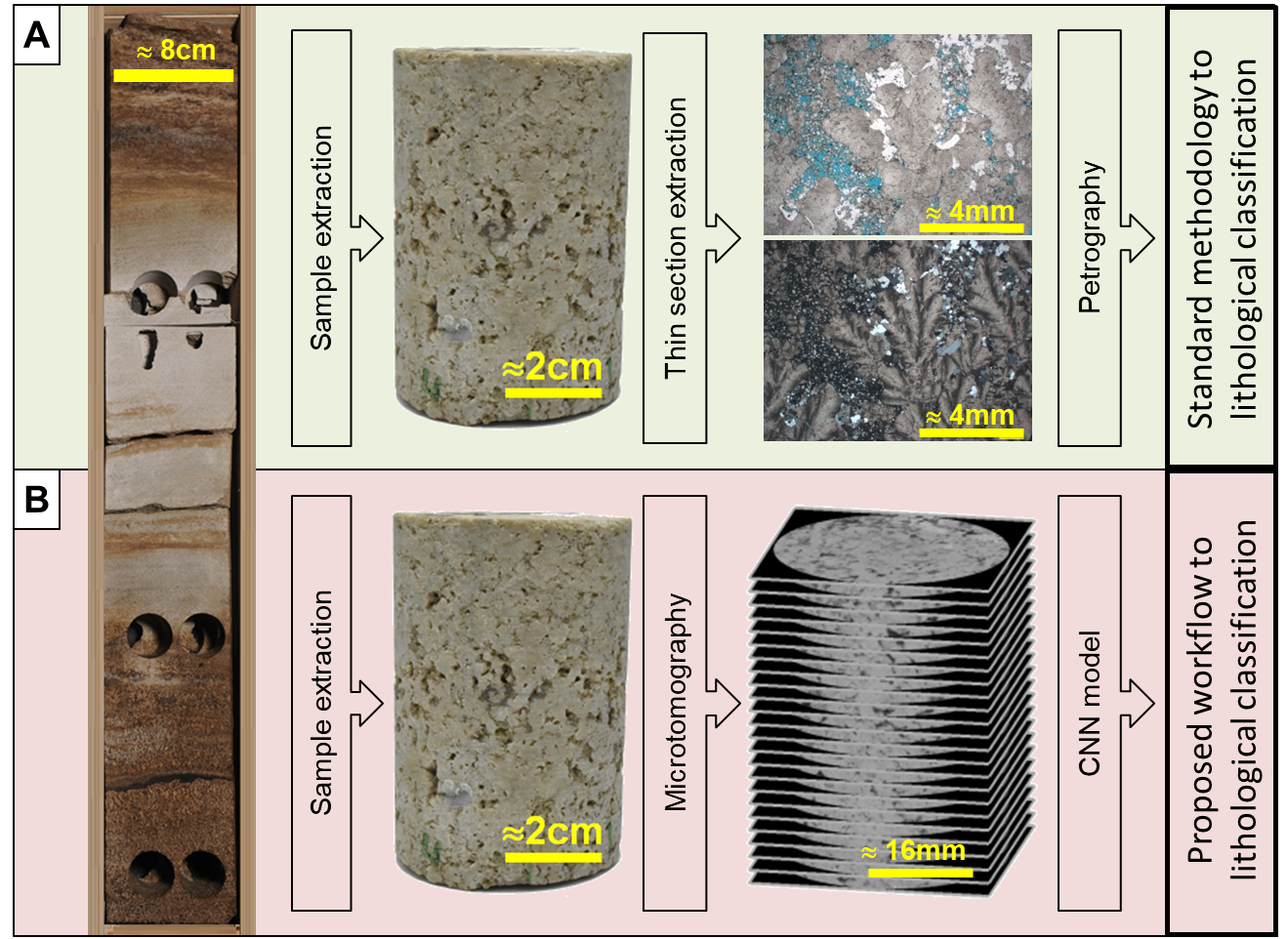}
    \caption{\textbf{a} Standard and \textbf{b} proposed methodologies for lithological classification}
    \label{fig:standardandproposed_workflow}
\end{figure*}

The standard method for lithology classification is supported by individual thin-section image analysis, a time-consuming process that leads to subjective interpretations of rocks. In Fig. \ref{fig:standardandproposed_workflow}a, the usual steps for the lithological definition, starting with sample extraction and posterior thin-section analysis, are shown. The thin-section images are strictly bidimensional, cover a small area and are directly correlated with the orientation extraction besides being destructive.

In Fig. \ref{fig:standardandproposed_workflow}b, the suggested workflow to classify carbonate sample lithologies based on micro-CT images is shown. X-ray micro-CT is a relatively fast, non-destructive, and inexpensive technique that can be incorporated into the reservoir characterization workflow before any destructive testing, thus producing images that closely correspond to serial thin-sections through a sample. For more details, a complete overview of image acquisition and micro-CT reconstruction of earth materials can be found in Cnnudde and Boone \cite{cnuddeandboone} and Hanna and Ketcham \cite{hannaandketchan}. The data can be useful for visualization tasks and quantitative and qualitative characterization of the internal features, delimited by variations in density and atomic composition. Another advantage of the recommended procedure is the capability of using a more representative portion of rock samples. 


\section{Convolutional neural networks}

One of the most used architectures for vision tasks is the CNN. It was initially proposed by LeCun et al. \cite{lecun1989generalization} but had its popularity increased when used by Krizhevsky et al. \cite{krizhevsky2012imagenet} in the \textsc{ImageNet} contest. The main reasons for its popularity are the superior performance compared with conventional approaches used in image analysis, along with efficient implementation on GPUs.

\begin{figure*}[!ht]
\centering
\includegraphics[scale = 0.8]{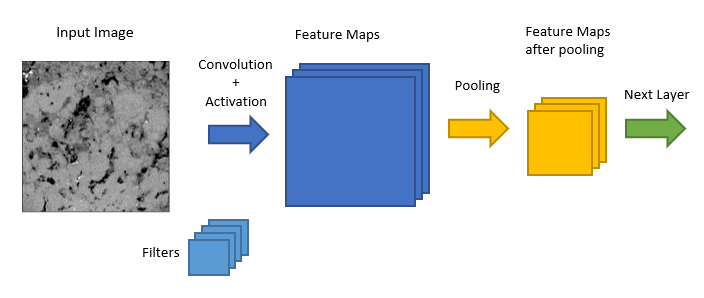}
\caption{Example of the combination of convolutional, activation, and pooling layers using a grayscale image as input}
\label{fig:convStructure}
\end{figure*}

The core element of this network, in simple form, is the sequence of three consecutive layers: convolutional, activation, and pooling layers. This sequence of operations can be sequentially stacked as many times as desired. This aspect of the model is usually known as feature extraction. It is expected that models with more layers will be able to extract more complex features because the receptive field of the network increases as the input passes forward through each layer, thus allowing the interpretation of spatial structures on different scales. After this process, the extracted features are provided as input for a fully connected network that accomplishes the mapping between the features and the desired output. The overall structure of this network is shown in Fig. \ref{fig:convStructure}. Details of these elements are provided below.

\begin{enumerate}
\item{\textbf{Convolutional layer}}

The convolutional layer consists of a set of trainable filters which have a receptive field that depends on the filter size. Each filter is convolved across the dimensions of the input, thus computing the dot product between the filter parameters and the corresponding input section. As a result of this operation, a feature map is obtained. It is important to note that the filter parameters are shared across all input images, therefore reducing the total amount of parameters when compared with a fully connected layer. This operation can be represented as
\begin{equation}
\label{eq:convoperation}
\mathbf{o}_{l} = f((\mathbf{o}_{l-1} * \mathbf{W}_{l}) + \mathbf{b}_{l}),
\end{equation}
where $\mathbf{o}_{l}$ is the output tensor of the $l$th layer, $\mathbf{o}_{l-1}$ is the previous layer output or the raw input image when $l=1$, $*$ is the convolution operator, $\mathbf{W}_{l}$ is the weights tensor, $\mathbf{b}_{l}$ is the bias vector and $f$ is an activation function. This operation allows preservation of spatial information and the analysis of multidimensional structure, which is desirable in computer vision problems to detect textures, shapes, complex structures, and other spatial features \cite{Goodfellow}.

The activation function applied after the convolution operator is used to insert linear or nonlinear behavior, as required, into the model mapping function. Currently, the rectified linear unit (ReLU) is one of the most commonly utilized methods to accomplish this; however, others such as hyperbolic tangent or sigmoid function can also be used. 

\item{\textbf{Pooling layer}}

The pooling layer is used to decrease the spatial dimension of the inputs with the aim of reducing the computational cost of the model. Additionally, this operation provides for more robust feature extraction which is invariant with respect to small translations \cite{Goodfellow}. This operator is analogous to the convolutional operator. However, this layer executes a selection among the elements of the window without trainable parameters. Two of the most common functions used for this type of layer are maximum and mean functions. 

\item{\textbf{Fully connected layer}}

For an image classification problem, it is usually necessary to add a fully connected layer to the neural network after the convolutional structure to enable class predictions \cite{krizhevsky2012imagenet,Goodfellow}. This fully connected layer is added to perform the mapping from feature space to domain specific space. Mathematically, it is possible to define this layer as
\begin{equation}
\label{eq:fcoperation}
\mathbf{h}_{l} = f(\mathbf{h}_{l-1} \mathbf{W}_{l} + \mathbf{b}_{l}),
\end{equation}
where $\mathbf{h}_{l}$ is the output vector, $\mathbf{h}_{l-1}$ is a flattened version of the $\mathbf{o}_{l-1}$ tensor, $\mathbf{W}_{l}$ represents the $l$th layer weights tensor, $\mathbf{b}_{l}$ is the bias vector of layer $l$ and $f$ is the activation function.
\end{enumerate}

The simplified version of the CNN model for image classification, as previously described, can imply some problems and limitations of the model capabilities with emphasis on risk of model overfitting. The literature presents different methods to overcome this problem, ranging from data augmentation to regularization techniques, among others. In the line of regularization techniques, the dropout layer \cite{JMLR:v15:srivastava14a} has been proposed.  This layer is an inexpensive computational operator that consists of setting a percentage of layer outputs to zero while training is executed. Dropout training is compared to training an ensemble of models that share the model parameters. Thus, this layer has demonstrated improvements of the neural network generalization ability and prevented model overfitting \cite{krizhevsky2012imagenet}.

Another problem associated with the use of the fully connected layer directly after the last layer of the feature extraction phase is the number of parameters. This layer is accountable for most of the model parameters and, consequently, for most of the computational cost, since each element from the input is connected to all neurons in the layer. This dense connection also limits the input dimension to a fixed value after model construction, which is undesirable in some applications.  To overcome these problems, the global average pooling (GAP) layer, consisting of applying an average pooling operator with the same size of the input, thus generating a unique value for each feature map, has been proposed by Lin et al. \cite{lin2013network}. This procedure allows the use of input images with different dimensions, since the output of the GAP layer corresponds to the number of feature maps that this layer receives as input.

\begin{figure*}[!ht]
\centering
\includegraphics[scale=0.8]{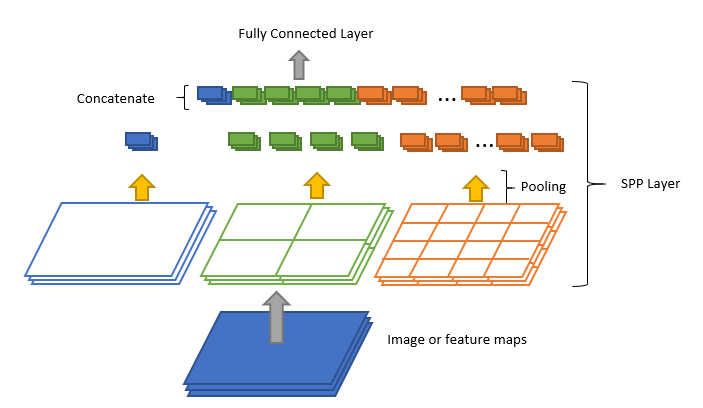}
\caption{SPP example with three pyramid levels}
\label{fig:SPPLayer}
\end{figure*}

Adding a GAP layer greatly reduces the number of trainable parameters. Nevertheless, the average of each feature map may not be sufficiently representative. Multiscale analysis may be more representative and achieve better results, as was conducted in He et al. \cite{heetal}, where the inclusion of a spatial pyramid pooling (SPP) layer \cite{1641019} in a CNN is proposed. The SPP consists of applying different pooling operators with filter sizes that depend on the input dimensions and the number of pyramid levels to be used, as is shown in Fig. \ref{fig:SPPLayer}.

The number of pyramid levels in the SPP is arbitrary. With more levels, more splits of the input data will be generated. In the scenario presented in Fig. \ref{fig:SPPLayer}, three pyramid levels are used, which split an image in three ways. The first level applies the pooling on the entire image, the second divides it into four parts, applying the pooling to each part, and, finally, the third level divides the image into sixteen parts, applying pooling to each one of the parts. In this case, a $21(1+4+16)$-dimensional vector is generated for each image or feature map, therefore also allowing the use of differently sized images as input to the network, since it generates a vector of the same size for any given input.

In this work, a simplified version of CNN is used as a base model, which consists of a few CNN structures followed by fully connected layers. From this base model, three variations are tested: adding a SPP layer and/or GAP.


\section{Materials and methods}

\subsection{Dataset}

For training and testing the model, 60 plug samples extracted from a carbonate reservoir located in a Brazilian pre-salt oilfield were used. The micro-CT images of each sample were cropped to remove the void area around each rock sample.

Supported by thin-section photomicrographs, the 60 samples, extracted from the same depth, were classified by geologists into three categories - grainstones, spherulites, and stromatolites. This imbalance is due to the reservoir characteristics at well depth and location. The micro-CT images and lithological labels correspond to the input-target pairs for supervised model training.

The image resolution ranged from 26.8 $\mu m$ to 50.5 $\mu m$, primarily due to sample size and acquisition system arrangement. Given the high resolution of micro-CT images, which result in almost identical consecutive set images, 100 equally spaced slices from each sample were selected to avoid using images which were excessively similar. Assuming each plug as homogeneous, all slices from the same sample received the same lithological label. The dataset is distributed of 18 grainstones, 18 spherulites, and 24 stromatolites samples. Therefore, the dataset was composed of 6,000 images: 1,800, 1,800, and 2,400 for each class, respectively. 

This dataset presents some features that may cause misclassification problems, such as ambiguity in the definition of the lithology. Usually, grainstones consist of other lithology fragments, which can generate errors in the training step, for example. Additionally, one can underline details observed in thin-section images, which cannot be appreciated in micro-CT images due to resolution constraints. Also, different luminous conditions are produced due to the polychromatic nature of the X-ray beam. At last, image artifacts can be caused by high-density materials.

To save time and prevent specific impacts on automated classification, the images were used without any filtering step or artifact removal, which is quite common in several related works. Accordingly, some preprocessing steps were applied to circumvent the problem of different luminous conditions and the problem of processing images of different sizes. Additionally, grayscale images were used to avoid the need for segmentation computation and possible associated user-bias.

The preprocessing step consists of scaling each image according to its mean and standard deviation. In addition to that, in order to enable the use of a typical CNN model, it is necessary to reduce the dimensions of the images to some value. The technique used for this is known as resizing: this function applies a bilinear interpolation of pixels to reduce the image to a specified size. For this purpose, the \textsc{OpenCV Python} package was used, and the fixed output size was set to $256 \times 256$.

\subsection{Models}

In this work, a CNN model is used to classify lithology of micro-CT images of plugs. On top of this base model, a few modifications were made before the first fully connected layer, thus resulting in four different topologies, which have their configurations represented in Table \ref{table:AllModels}.

\begin{enumerate}
\item{ \textbf{Model 1}}

The first model tested to solve this classification problem was a CNN structure directly followed by fully connected layers, with input data restricted to a fixed size. As shown in Table \ref{table:AllModels}, this model has three sequential convolutional structures that consist of a convolution layer with filters of $3 \times 3$ followed by ReLU as activation function and a max-pooling layer. The convolutional operations are performed with padding of 1 and stride of 1, where the first term refers to the addition of a border on the input and the latter term refers to the number of pixels skipped by the convolution operator. The convolutional layers have 64, 48 and 32 filters, respectively. These convolutional structures are followed by a fully connected layer with activation ReLU and 200 neurons, a dropout layer with 50\% chance to drop connections, and the output layer.

\item{\textbf{Model 2}}

The second model replaces the last pooling layer of the previous model with a SPP layer, thus reducing the number of parameters and enabling the use of images with different sizes.

\item{\textbf{Model 3}}

In this case, a GAP layer is used as an alternative to the SPP layer, thus allowing reduction of the number of parameters compared with the two previous models at the cost of losing multiscale representation when compared with Model 2.

\item{\textbf{Model 4}}

Aiming to take advantage of Model 2 and Model 3, multiscale analysis and parameter reduction, respectively, Model 4 consists of simultaneously using two operations, i.e., the SPP layer followed by a GAP layer, as an alternative to the last pooling layer of Model 1.

\begin{table}[ht]
\caption{Model configurations.}
\begin{tabular}{|c|c|c|c|}
\hline
Model 1                   & Model 2                                  & Model 3                                 & Model 4                 \\ \hline
\multicolumn{4}{|c|}{Convolutional Layer 3x3 - 64 filters}                                                                               \\ \hline
\multicolumn{4}{|c|}{ReLU}                                                                                                               \\ \hline
\multicolumn{4}{|c|}{Max Pooling}                                                                                                           \\ \hline
\multicolumn{4}{|c|}{Convolutional Layer 3x3 - 48 filters}                                                                               \\ \hline
\multicolumn{4}{|c|}{ReLU}                                                                                                               \\ \hline
\multicolumn{4}{|c|}{Max Pooling}                                                                                                           \\ \hline
\multicolumn{4}{|c|}{Convolutional Layer 3x3 - 32 filters}                                                                               \\ \hline
\multicolumn{4}{|c|}{ReLU}                                                                                                               \\ \hline
\multirow{2}{*}{Max Pooling} & \multirow{2}{*}{SPP} & \multirow{2}{*}{GAP} & SPP \\ \cline{4-4} 
                          &                                          &                                         & GAP  \\ \hline
\multicolumn{4}{|c|}{Fully Connected Layer - 200 neurons}                                                                                \\ \hline
\multicolumn{4}{|c|}{ReLU}                                                                                                               \\ \hline
\multicolumn{4}{|c|}{Dropout}                                                                                                            \\ \hline
\multicolumn{4}{|c|}{Fully Connected Layer - Softmax}                                                                                    \\ \hline
\end{tabular}
\label{table:AllModels}
\end{table}

\end{enumerate}

The number of trainable parameters of each model is shown in Table \ref{tab:modelsParamsComparision}, which clarifies one of the main differences between all topologies used. Model 1 is almost 38 times larger than Model 2 and 136 times larger than Models 3 and 4, which means that there is a considerable amount of parameters to train using the same data.

\begin{table}[!ht]
\centering
\caption{Comparison of the parameters of the models.}
\label{tab:modelsParamsComparision} 
\begin{tabularx}{\textwidth}{Xl}
\hline\noalign{\smallskip}
Model & Parameters \\
\noalign{\smallskip}\hline\noalign{\smallskip}
    Model 1 &    6,596,595 \\
    Model 2 &    177,395 \\
    Model 3 &    49,395 \\
    Model 4 &    49,395 \\
\noalign{\smallskip}\hline
\end{tabularx}
\end{table}

\subsection{Training}

All models used in this work have trainable weights and biases, which can be represented as a set of matrixes as follows
\begin{equation}
\label{eq:weightsSet}
    \theta = \{ \mathbf{W}_{1}, \mathbf{W}_{2}, ..., \mathbf{W}_{L}, \mathbf{b}_{1}, \mathbf{b}_{2}, ..., \mathbf{b}_{L} \},
\end{equation}
where $L$ is the number of layers, $\mathbf{W}_{l}$ ($l=1, 2, ..., L$) denotes the weights of each layer and  $\mathbf{b}_{l}$ ($l=1,2,...,L$) their respective biases.

To achieve the best solution, it is necessary to update those weights, which is performed by minimizing the loss between the predicted and real values. For this work, the cross-entropy loss function was chosen to be used by all models implemented, which can be formulated as
\begin{equation}
\textit{L}(y,\hat{y};\theta) =  - \frac{1}{N} \sum^{C}_{c} \sum^{N}_{n}  y_{c,n}log(\hat{y}_{c,n}(\theta)),
\end{equation}
where $N$ is the amount of samples, $C$ is the number of classes, $y_{c,n}$ is the target value of class $c$ of the nth sample and $\hat{y}_{c,n}$ is the neural network predicted probability of class $c$ of the nth sample. The loss function is minimized using the back-propagation algorithm proposed by Lecun et al. \cite{lecun1998gradient} and optimized using the Adam algorithm \cite{kingmaandba} with $0.001$ learning rate, $\beta_0=0.9$ and $\beta_1=0.999$. When optimizing a neural network model, it is also necessary to use a criterion to stop that optimization: thus, the validation set was used for that purpose.

Moreover, it is worth noticing that due to the imbalance of classes in the dataset, the gradient weighting technique was used \cite{Goodfellow}. The ponderation was inversely proportional to the occurrence of the classes, in which the lower frequency classes were multiplied by a constant greater than one in order to increase the gradient value. In the present case, the value is 1.33, while the higher frequency class had its gradient multiplied by one. Thus, the gradient applied to the neural network weights is balanced, given that if this technique were not used, the trained model would be biased by the class with the highest occurrence within the dataset.

The infrastructure used for this work was one \textsc{NVIDIA TITAN V}, granted by \textsc{NVIDIA}, and the models trained with original images required approximately 90 seconds per epoch, while those trained on resized images required approximately 30 seconds.

\subsection{Validation statistics}

There are many ways to evaluate a machine learning algorithm \cite{Japkowicz,friedman2001elements}. The selection of the appropriate evaluation functions depends on the task executed by the model, classification or regression, and characteristics of the problem, such as data imbalance. In this work, accuracy, recall, precision and $F_1$ score were used as validation statistics. Their definitions are listed below.

\begin{enumerate}
\item \textbf{Accuracy}

Accuracy is the measure of how many elements were correctly classified over all samples.

\item \textbf{Precision}

Precision indicates the amount of samples of one class correctly predicted over the amount of samples predicted as that class. Its formulation is given by
\begin{equation}
Precision (C_{i}) = \frac{TP_i}{TP_i+FP_i},
\end{equation}
where $TP_i$ or True Positive is the number of correctly classified samples of class $C_i$ and $FP_i$ is the number of samples from the other classes which were classified as class $C_i$. Thus, the denominator represents the amount of elements predicted as class $C_i$.

\item \textbf{Recall}

Recall is the amount of correctly predicted samples of one class over the amount of samples of that class. Its formulation is given by

\begin{equation}
Recall (C_{i}) = \frac{TP_i}{TP_i+FN_i},
\end{equation}
where $TP_i$ is the number of samples correctly predicted, $FN_i$ is the number of samples of class i that are misclassified, and $TP_i+FN_i$ is the total number of samples of class $C_i$.

\item \textbf{$F_{1}$ score}

$F_{1}$ score is a measure that considers both recall and precision of one class to be calculated. Its formulation is given by
\begin{equation}
F_{1}\ Score(C_{i}) = 2\times\frac{Precision_{i}\times Recall_{i}}{Precision_{i}+Recall_{i}},
\end{equation}
where $i$ represents the class. The numerator is the product of precision and recall of class $C_i$, and the denominator is the sum of precision and recall of class $C_i$.

\end{enumerate}

\subsection{Nested k-folds}

The cross-validation method is one of the most popular for prediction error estimation in machine learning problems \cite{Japkowicz,friedman2001elements} through estimation of variance among different testing sample results. This method consists of splitting the dataset into $k$ subsets of equal or nearly equal sizes, where each subset is known as a fold and is stratified: that is, each fold attempts to retain the same class distribution. A model is then trained with $k-1$ folds and tested on the remaining fold, and this process is repeated $k$ times until all sets have been used for testing once. Thus, the experiment returns $k$ estimates of the model classification error.

The nested k-fold method used in this paper consists of applying this approach by adding a validation set and first splitting the dataset into $k$ folds, where $k-1$ folds are used for training and validation, and the remaining fold for testing: within the training and validation set, one fold is separated for validation, with the rest used for training.

A model is then trained using the validation set as the stopping criterion after the training is completed, another model is instantiated, and the validation fold is changed. This process is repeated until all folds have been used for validation once. In this way, we train $k-1$ models for each test set, and then the test set is changed and the process repeated, thus training $k\times(k-1)$ models in total.

\begin{figure}[!ht]
    \centering
    \includegraphics[scale=0.388]{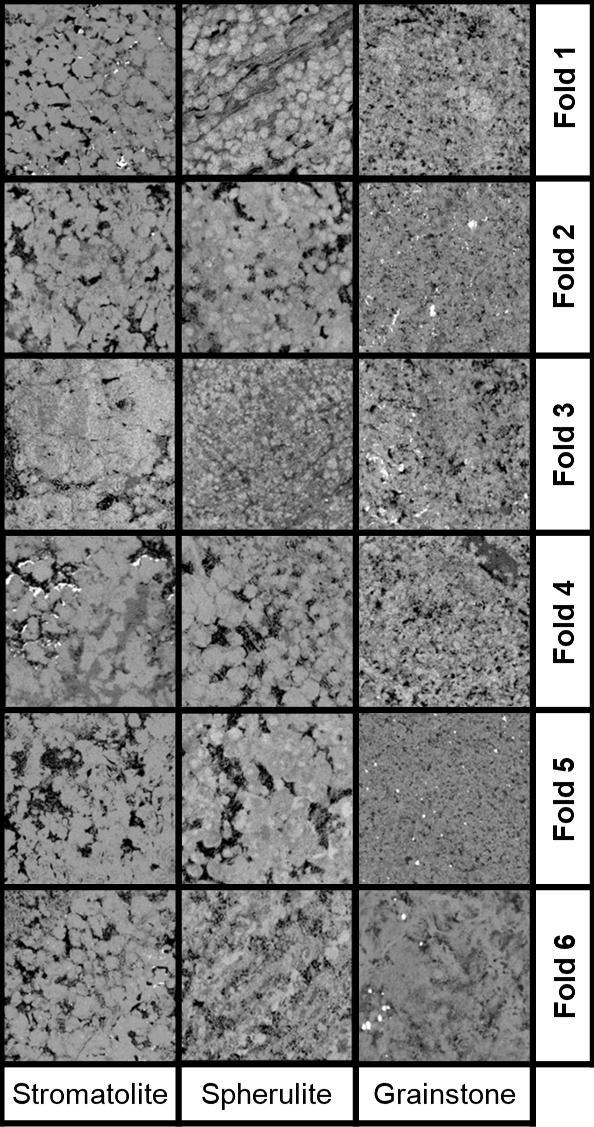}
    \caption{Examples of each lithology class composing each fold}
    \label{fig:6k-folds}
\end{figure}

For this work, we used six fixed folds of plugs for nested cross-validation. Examples of each class of each fold can be observed in Fig. \ref{fig:6k-folds}. To avoid adding more bias to the obtained results, the splitting of folds was performed based on plugs instead of images, given that we assumed the same class for all plug images, as stated in a previous section. All folds had the same class distribution: three spherulites, four stromatolites, and three grainstones.

\section{Results}

The tested scenarios were chosen in order to compare the differences in the performances of models using original (Models 2, 3, and 4) and resized (Models 1, 2, 3 and 4) images. Additionally, we sought to verify whether adding a SPP and/or GAP layer would improve the performance of the trained models. There are two different approaches to evaluate our models' performance. The first one is to individually classify each image, while the second takes into account the most frequent classification of all images of a given plug. The results are summarized hereafter as tables, where the values express the mean and the standard deviation of the nested k-fold trained model performance.

\subsection{Evaluation per image}

Results available in Tables \ref{tab:AccResultsPerimage} and \ref{tab:F1ScoreTestResultsPerimage} show the model performance when individually evaluating each image of the test set. The first Tables \ref{tab:AccResultsPerimage} shows the overall accuracy, while the Table \ref{tab:F1ScoreTestResultsPerimage} presents F1 scores. 

The models that achieved the best mean accuracy, shown in Table \ref{tab:AccResultsPerimage}, were trained on resized images and used both SPP and GAP techniques. However, almost all models are within the same error bar of one another. This large error bar probably occurred due to the heterogeneity of carbonate rocks and the small number of samples available, thus generating a dependence upon the samples used for model training. We expect that with more samples, the models could be more effectively trained for generalization, since we would therefore employ a more representative dataset.

\begin{table}[!ht]
\centering
\caption{Mean and standard deviation of accuracy over the nested k-fold results evaluating each image individually.}
\label{tab:AccResultsPerimage} 
\begin{tabularx}{\textwidth}{XXl}
\hline\noalign{\smallskip}
Image type & Model &  Test (\%) \\
\noalign{\smallskip}\hline\noalign{\smallskip}
    Original & Model 2 & $63.96\pm10.79$  \\
    Original & Model 3 & $65.19\pm12.18$ \\
    Original & Model 4 & $70.21\pm11.90$ \\
    Resized & Model 1 & $61.67\pm07.29$ \\
    Resized & Model 2 & $75.54\pm08.68$ \\
    Resized & Model 3 & $73.26\pm09.87$  \\
    Resized & Model 4 & $76.62\pm10.50$ \\
\noalign{\smallskip}\hline
\end{tabularx}
\end{table}

Analyzing Table \ref{tab:F1ScoreTestResultsPerimage}, the models trained on resized images exhibited better mean performance for most of the classes. These models also exhibited better mean performance with respect to the grainstone class. Using this evaluation approach, the authors found the best model using resized images to be Model 2; that choice was made due to smaller standard deviation and high mean with respect to both accuracy and F1 scores.

\begin{table*}[!ht]
\centering
\caption{Mean and standard deviation of F1 score over the nested k-fold results evaluating each image individually.}
\label{tab:F1ScoreTestResultsPerimage}
\begin{tabularx}{\textwidth}{XXXXl}
\hline\noalign{\smallskip}
Image type & Model & Grainstone (\%) & Spherulite (\%) & Stromatolite (\%) \\
\noalign{\smallskip}\hline\noalign{\smallskip}
    Original & Model 2 &   $60.73\pm15.34$ & $59.44\pm15.57$ &   $69.01\pm12.51$ \\  
    Original & Model 3 &   $65.18\pm15.06$ & $60.90\pm17.82$ &   $66.73\pm14.62$ \\  
    Original & Model 4 &   $68.18\pm15.98$ & $66.28\pm17.74$ &   $73.59\pm14.07$ \\ 
    Resized & Model 1 &   $65.59\pm10.03$ &  $48.44\pm09.90$ &   $65.64\pm10.95$ \\ 
    Resized & Model 2 &   $78.51\pm11.51$ &  $71.20\pm10.77$ &   $75.45\pm12.71$ \\ 
    Resized & Model 3 &   $77.77\pm10.30$ &  $64.49\pm17.97$ &   $74.68\pm13.03$ \\
    Resized & Model 4 &   $78.90\pm11.48$ &  $74.97\pm10.07$ &   $74.99\pm15.19$ \\
\noalign{\smallskip}\hline
\end{tabularx}
\end{table*}

\subsection{Evaluation per plug}

Results of the most frequent classes of plug image predictions are shown in Tables \ref{tab:AccResultsPerPlug} and \ref{tab:F1ScoreTestResultsPerPlug}. This evaluation approach is motivated by our purpose to classify the full sample. Assuming that the plugs are homogeneous, we seek to obtain a unique label for each plug sample. 

Table \ref{tab:AccResultsPerPlug} presents the overall accuracy for each set, where is possible to verify that the mean accuracy increased for all models, but the standard deviation increased as well. This occurred because there are just 10 plugs in each of those sets, and just one mislabeled plug thus represents 10\% accuracy.
\begin{table}[h]
\centering
\caption{Mean and standard deviation of accuracy over the nested k-fold results evaluating each plug.}
\label{tab:AccResultsPerPlug} 
\begin{tabularx}{\textwidth}{XXl}
\hline\noalign{\smallskip}
Image type & Model &  Test(\%) \\
\noalign{\smallskip}\hline\noalign{\smallskip}
    Original & Model 2 & $67.33\pm13.63$ \\
    Original & Model 3 & $68.00\pm14.95$ \\
    Original & Model 4 & $73.67\pm14.50$ \\
    Resized & Model 1 & $70.33\pm15.20$ \\
    Resized & Model 2 & $81.33\pm11.06$ \\
    Resized & Model 3 & $76.33\pm10.98$\\
    Resized & Model 4 & $78.00\pm14.48$ \\
\noalign{\smallskip}\hline
\end{tabularx}
\end{table}

For F1 scores, Table \ref{tab:F1ScoreTestResultsPerPlug} shows that most models also exhibited the same behavior of an increase in both mean and standard deviation for each class, which means that on average the models were able to identify most of each plug correctly for all classes. The model selected by the authors based on these results was also Model 2 for resized images, since it still exhibits high mean and low standard deviation compared to the other models with respect to both accuracy and F1 score.


\section{Discussion}

According to accuracy and F1 score, the best fold of the validation set was selected from Model 2 trained on resized image for deeper analysis. Table \ref{tab:RecallResultsPerClass} shows the model performance when evaluating each image individually, along with the plug classifications for each set. The sets are divided as follows: approximately $4/6$ of all images comprised the training set, $1/6$ the validation set and the remaining $1/6$ the test set. The class distribution of each set was $3/10$ grainstones, $4/10$ stromatolite and $3/10$ spherulite.

Evaluating the results by lithological classes, the classification of the grainstone class achieved the best results, besides the feasibility of grainstones can be confused with other lithologies on account of this lithology being formed by a large variability of fragments. These fragments are the base structures of different types of carbonate rocks, for example stromatolites, and their predominance can lead to misclassification.

\begin{figure*}[!ht]
    \centering
    \includegraphics[height=3.05cm]{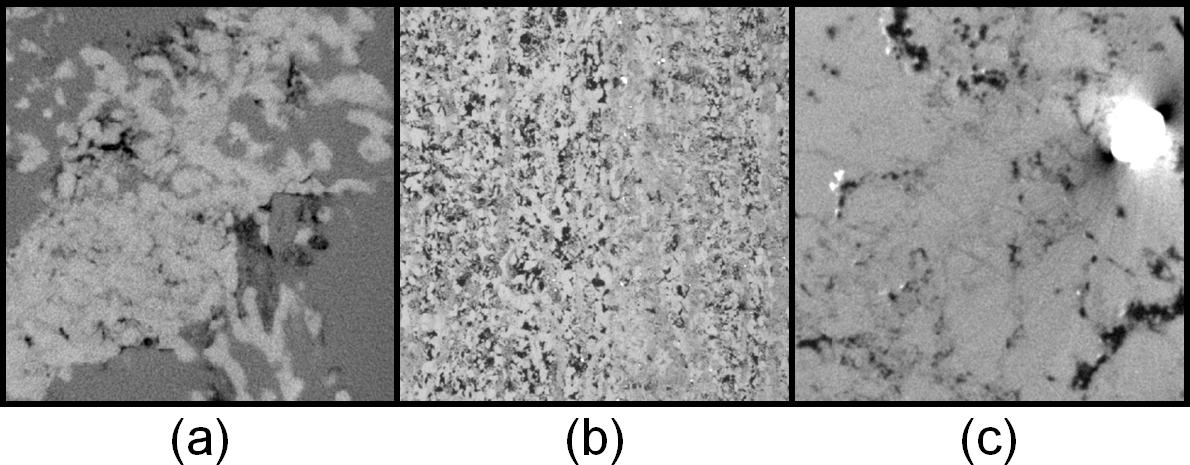}
    \caption{Possible features that lead some images to be classified as the wrong lithologies: \textbf{a} sample heterogeneity; \textbf{b} dissolution process; and \textbf{c} high-density artifacts}
    \label{fig:erros_grs_est}
\end{figure*}

From the entire group of 18 grainstone samples, a total of 17 were correctly classified with 100\% accuracy. The remaining sample, included in the test fold, was also classified as grainstone. However, 3\% of these images were erroneously labeled, which was primarily due to sample heterogeneity, comprising fragments of spherulites and stromatolites, in addition to intense recrystallization and the replacement of calcite cement by dolomite and quartz, as noticed in Fig. \ref{fig:erros_grs_est}a.

The stromatolite class also presented satisfactory results. All 24 stromatolite samples were properly categorized, with an average of approximately 95\% of the accuracy of image classification. Considering the training set, only one of the samples represented 8\% of the images mislabeled as grainstone, possibly as a result of dissolution and the small size of the features of the images, as shown in Fig. \ref{fig:erros_grs_est}b. It is important to highlight that those were the only errors of the training set images. 

Regarding the validation set, two of four samples presented misclassified images which were labeled as spherulite. Those represent less than 0.5\% of the stromatolite class images. Through image-by-image analysis, 6 of 10 of these images present artifacts caused by high-density minerals, exhibiting star patterns in images, as can be noticed in Fig. \ref{fig:erros_grs_est}c. At first glance, this seems to be the main reason for algorithm deficiency.

\begin{table*}
\centering
\caption{Mean and standard deviation of accuracy over the nested k-fold results evaluating each plug.}
\label{tab:F1ScoreTestResultsPerPlug} 
\begin{tabularx}{\textwidth}{XXXXl}
\hline\noalign{\smallskip}
Image type & Model & Grainstone (\%) & Spherulite (\%) & Stromatolite (\%) \\
\noalign{\smallskip}\hline\noalign{\smallskip}
    Original & Model 2 &  $63.89\pm20.16$ &  $63.83\pm24.99$ &   $70.31\pm16.86$ \\   
    Original & Model 3 &  $67.26\pm19.90$ &  $64.92\pm22.90$ &   $68.26\pm17.06$ \\   
    Original & Model 4 &  $71.90\pm20.61$ &  $67.33\pm28.35$ &   $76.55\pm15.60$  \\  
    Resized & Model 1 &   $76.16\pm16.37$ &  $47.56\pm29.01$ &   $75.09\pm17.40$  \\  
    Resized & Model 2 &   $83.51\pm15.37$ &  $78.75\pm14.62$ &   $79.99\pm15.00$ \\ 
    Resized & Model 3 &   $78.25\pm12.80$ &  $67.48\pm22.90$ &   $79.01\pm13.70$ \\
    Resized & Model 4 &   $78.03\pm18.22$ &  $78.26\pm15.52$ &   $75.89\pm18.08$\\
\noalign{\smallskip}\hline
\end{tabularx}
\end{table*}

\begin{table*}
\justify
\caption{Recall of each class of best validation fold according to accuracy and mean F1 score from Model 2 trained on resized image models per image and per plug.}
\label{tab:RecallResultsPerClass} 
\begin{tabularx}{\textwidth}{XXXXl}
\hline\noalign{\smallskip}
Evaluation & Recall (\%) & Grainstone & Spherulite & Stromatolite \\
\noalign{\smallskip}\hline\noalign{\smallskip}
Image & Train & 100.00 & 100.00 & 99.50  \\
Image & Validation & 100.00 & 91.34 & 97.50  \\
Image & Test & 99.00 & 43.67 & 77.25  \\
Plug & Train & 100.00 & 100.00 & 100.00  \\
Plug & Validation & 100.00 & 100.00 & 100.00  \\
Plug & Test & 100.00 & 33.34 & 100.00  \\
\noalign{\smallskip}\hline
\end{tabularx}
\end{table*}

\begin{figure*}[!ht]
    \centering
    \includegraphics[height=3.05cm]{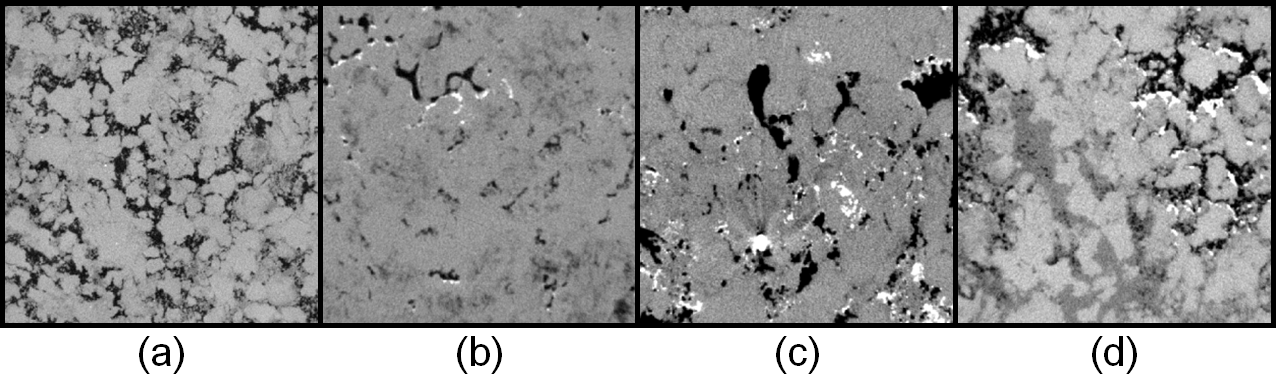}
    \caption{Increase in post-depositional processes resulting in high rates of misclassified stromatolite images, from \textbf{a} to \textbf{d}}
    \label{fig:erros_est}
\end{figure*}

Finally, for the test fold, 23\% of the images from four different samples were mislabeled as spherulite. The error rate increased according to the level of post-depositional processes affecting the stromatolite samples and mud invasion on the porous system. Each stromatolite sample of the test set presented in Fig. \ref{fig:erros_est}a-d presents 9\%, 11\%, 24\%, and 47\% of error, respectively.

\begin{figure*}[]
    \centering
    \includegraphics[height=3.05cm]{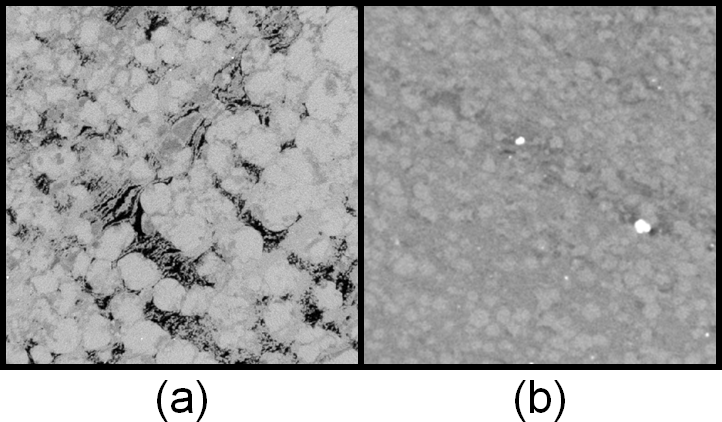}
    \caption{Spherulite test set samples misclassified as \textbf{a} stromatolite and \textbf{b} grainstone}
    \label{fig:erros_esf}
\end{figure*}

\begin{figure*}[!ht]
    \centering
    \includegraphics[height=3.05cm]{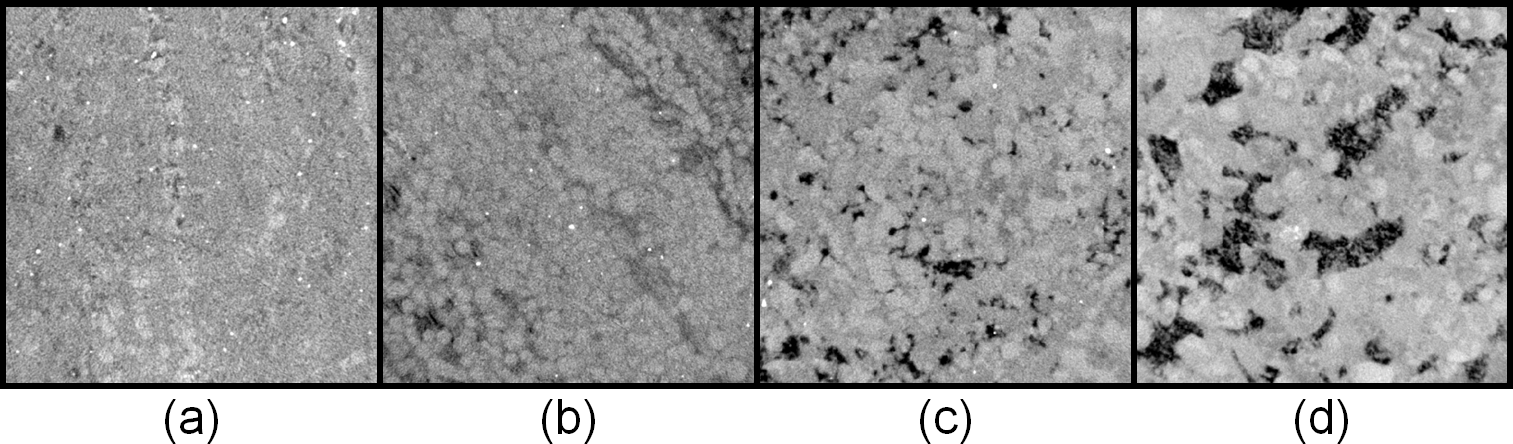}
    \caption{Increase in pore-size distribution of spherulite images from \textbf{a} to \textbf{d}}
    \label{fig:erros_esf2}
\end{figure*}

At last, in the spherulite class, two of three samples from the test set were confused. The first one, classified as stromatolite with 93\% precision, presents mud-filled laminated structures and high cementation and dolomitization levels, as exhibited in Fig. \ref{fig:erros_esf}a. On the other side, labeled as grainstone with 63\% accuracy, a more compact sample in Fig. \ref{fig:erros_esf}b exhibits interparticle porosity filled with dolomite filaments and mud. For the 15 remaining samples, just less than 3\% of the images were confused, all of them belonging to validation and test sets. It can be observed that this inaccuracy follows a pattern in which some images of pore-filled samples are generally classified as grainstone, while porous samples are tagged as stromatolites. Each sample displayed in Fig. \ref{fig:erros_esf2}a-c presented 1\%, 13\%, and 21\% of images tagged as grainstone, while Fig. \ref{fig:erros_esf2}d presented 3\% of images classified as stromatolites. None of the training set spherulite images were confused.


\section{Conclusion and future works}

Despite the popularity of deep learning and CNN models, applications of these robust techniques to digital rock analysis, especially to lithology classification of carbonate rock based on micro-CT images, are still in development.

In this paper, we proposed a workflow for lithological classification of Brazilian pre-salt carbonate rock samples. The new approach is based on deep learning algorithms using micro-CT images as input. The use of non-destructive imaging techniques is a useful method for quantitatively and qualitatively analyzing rock samples, thus accomplishing the identification of most of the common features used in lithology classification. The image dataset includes three common lithologies of carbonate rocks, with 60,000 micro-CT images in total. The experiments show that the technique utilized to categorize the micro-CT images achieves an average of 75.54\% accuracy on the test set when analyzing images individually and achieves an average of 81.33\% accuracy on the test set when collectively analyzing plug images. These results are satisfactory for the preliminary use of CNN to label micro-CT images, since this was the first effort to apply deep learning algorithms for lithological classification of Brazilian pre-salt carbonate rocks micro-CT images.

As stated in a previous section, Model 2 trained on resized images shows some of the best performances of both image and plug evaluation. The authors chose this model because its accuracy exhibited high mean over the folds and lower standard deviation when compared to other models. Additionally, when evaluating the F1 score of each class of the test set, the model retained a high score for all classes.

The results obtained also made it possible to infer that using the resize preprocess and adding a SPP and/or GAP layer are better than not using them in the case of this dataset. Another reason to use resize preprocessing is the time consumption for model training and inferring the predictions once the resize models were approximately three times faster than the original image models. However, the preprocessing step can be enhanced and adapted to the dataset in order to mitigate some acquisition artifacts that could contribute to aggregate errors of the training steps.

This methodology could improve the speed of repetitive tasks to classify reservoir lithology, such as describing thin-sections, thus enabling the expert to focus on the most sophisticated tasks that require human intervention. In addition, the same produced data can be applied to other tasks, such as fluid flow and mechanical property estimations, with no need for computation of physical processes or laboratory tests.

Possible future enhancements of this work will focus on the use of additional samples for model training and evaluation and expansion of the number of lithological classes, testing other models, and transfer learning algorithms. Increasing the dataset may provide the ability to train models to recognize the most common post-depositional features, since dataset size proves to be one of the main factors for misclassifying some images. However, tasks of this complexity would require a huge training set. Another exciting work would be to predict the petrophysical properties of each plug and segment regions of interest.

\begin{acknowledgements}
The authors thank Petrobras for providing the data and financial support and NVIDIA Corporation for the GPU provided by the NVIDIA Grant Program. The authors also thank the Brazilian Research Council (CNPq) for the scholarships for students and researchers. 
\end{acknowledgements}

\bibliographystyle{spmpsci}      
\bibliography{bibliography}   

\end{document}